\documentclass[conference]{IEEEtran}
\IEEEoverridecommandlockouts
\usepackage{cite}
\usepackage{amsmath,amssymb,amsfonts}
\usepackage{graphicx}
\usepackage{textcomp}
\usepackage{xcolor}
\usepackage{enumitem}
\usepackage{booktabs}

\def\BibTeX{{\rm B\kern-.05em{\sc i\kern-.025em b}\kern-.08em
    T\kern-.1667em\lower.7ex\hbox{E}\kern-.125emX}}

\begin{document}


\title{FactoryLLM: A Safe and Open-Source AI Playground for Evaluating LLMs in Smart Factories \\}

\author{\IEEEauthorblockN{Yash Pulse, Yong-Bin Kang, Abhik Banerjee, Abdur Forkan, Prem Prakash Jayaraman}
\IEEEauthorblockA{{Swinburne University of Technology}, Melbourne, Australia \\
\{ypulse, ykang, abanerjee, fforkan, pjayaraman\}@swin.edu.au}
}

\maketitle

\begin{abstract}
Fault diagnostics and recovery in smart factories is challenging because critical information is dispersed across manuals of multiple machines which are interconnected through the manufacturing process. Large Language Models (LLMs) can provide a promising approach. In this paper, 
we propose \textit{FactoryLLM}, a safe and open-source AI playground designed for evaluating different LLM-based retrieval-augmented generation (RAG) models by analysing documents from multiple machines across the manufacturing process. 
FactoryLLM enables the user to configure the LLM, and assess performance when reasoning over multiple documents, through a dual evaluation setup using both RAGAS and NVIDIA’s ``LLM-as-a-Judge" metrics. FactoryLLM is safe because it allows users to run local or open-source LLMs without sharing sensitive industrial data, providing a controlled environment for experimentation. We demonstrate the efficacy of FactoryLLM through a case study which involves an Autonomous Intelligent Vehicle and its Mobile Planner software, evaluating three LLMs across 30 maintenance queries derived from approximately 600 pages of cross-machine documentation. The results suggest that FactoryLLM is effective in cross-machine document reasoning: every model achieved a groundedness score above 0.88. 
The full code and documentation for community to test FactoryLLM with their manufacturing specific scenarios are publicly available. \footnote{https://github.com/DigitalInnovationLab/Factory-LLM} 
\end{abstract}


\section{Introduction}

Modern factory floors increasingly operate as interconnected systems of systems rather than isolated collections of machines. Robotic arms, autonomous mobile robots, conveyor systems, programmable logic controllers, and supervisory software platforms work together to coordinate production workflows \cite{lu2017industry}, leading to ``self-healing factories". In such environments, faults rarely remain isolated. A failure in one component can propagate across multiple machines and software layers, leading to cascading disruptions throughout the production line. For example, an unexpected Autonomous Intelligent Vehicle (AIV) stoppage can stall downstream material handling, starve production cells, and force rapid diagnosis across interconnected systems. These cascading disruptions remain among the most costly operational challenges in smart manufacturing environments \cite{siemens2024true}.

Although industrial machines are accompanied by extensive technical documentation, diagnosing cross-system issues remains difficult. The key challenge is not the lack of information but the fragmentation of knowledge across multiple manuals written using different terminology and system abstractions. Operators often need to consult hundreds of pages of documentation across several machines to determine how a fault observed in one system may originate from another. For instance, a navigation failure reported by an AIV may actually stem from a configuration issue in its fleet management software rather than the vehicle hardware itself. Identifying such relationships requires connecting concepts across multiple documents and software layers.

Existing approaches provide limited ability to enable this type of cross-machine reasoning. Keyword-based search struggles when operator queries do not match the exact terminology used in manuals \cite{wan2014knowledge}. Rule-based expert systems require extensive manual knowledge engineering and are difficult to maintain as equipment evolves \cite{osti_5675197}. Knowledge graphs offer structured representations of machine relationships but become costly to construct and maintain as the number of machines increases \cite{zhang2025knowledge}. At the same time, the retirement of experienced technicians is widening the expertise gap, leaving fewer operators capable of diagnosing faults that span multiple systems \cite{siemens2024true}.

Large Language Models (LLMs) can provide a promising approach to address these challenges. Unlike rule-based or keyword-driven systems, LLMs can interpret natural-language queries, understand semantic relationships, and synthesise information from large collections of unstructured documents \cite{li2024large}. This capability enables them to act as an intermediary layer between operators and complex technical documentation. Retrieval-Augmented Generation (RAG) can further enhance this capability by grounding LLM responses in retrieved documentation rather than relying solely on the model’s training data \cite{lewis2020retrieval}. While several studies use LLM-based retrieval for documentation from a single machine, far fewer investigate their ability to reason across multiple machines and heterogeneous technical manuals. Cross-machine reasoning introduces additional challenges including aligning terminology across documents, retrieving relevant information from lengthy manuals, and synthesising evidence from multiple sources to explain a fault scenario.

To investigate these challenges, we propose \textit{FactoryLLM}, a safe and open-source LLM playground designed to evaluate how LLM-based RAG systems perform when reasoning across documentation from multiple industrial machines within a manufacturing process. Rather than serving solely as a troubleshooting assistant, FactoryLLM provides a controlled experimental environment where different LLM-based RAG systems—including local or open-source models that preserve privacy—can be plugged in and systematically assessed. The platform also provide robust assessment metrics of retrieval and generation capabilities under real-world manufacturing scenarios.
We demonstrate the suitability of FactoryLLM using a real-world case study stemming from  two tightly coupled systems in a smart factory: an \textit{AIV} and a \textit{Mobile Planner} software. The AIV is an autonomous mobile robot used to transport materials across factory floors, while the Mobile Planner is the fleet management software responsible for assigning missions, controlling navigation parameters, and coordinating multiple vehicles. Because the operational behaviour of the system emerges from the interaction between vehicle hardware and fleet management software, faults frequently manifest across both components. This makes the AIV–Mobile Planner ecosystem a suitable test case for evaluating cross-document reasoning. Using the technical manuals of both systems—each consisting of hundreds of pages of documentation, we demonstrate how FactoryLLM retrieves relevant information and synthesises grounded explanations that help diagnose issues spanning multiple machines. Through this case study, we show how FactoryLLM can support such cross-machine/document reasoning. Specifically, 
this paper makes the following contributions:

\begin{itemize}
\item We propose {FactoryLLM}, an LLM playground designed to support reasoning across multiple heterogeneous documentation spanning across machines and processes.
\item We demonstrate the efficacy of FactoryLLM through a realistic smart manufacturing case study involving an AIV robot and its Mobile Planner fleet management system to evaluate cross-machine troubleshooting scenarios.
\item We evaluate multiple LLMs and prompting strategies to assess cross-machine reasoning performance.
\end{itemize}

The rest of the paper is organised as follows. Section II reviews related work on LLMs and RAG in manufacturing. Section III describes the system architecture of FactoryLLM. Section IV presents the case study setup. Section V discusses results and evaluation. Section VI concludes the paper.

\begin{figure*}[t!]
    \centering
    \includegraphics[width=0.75\textwidth]{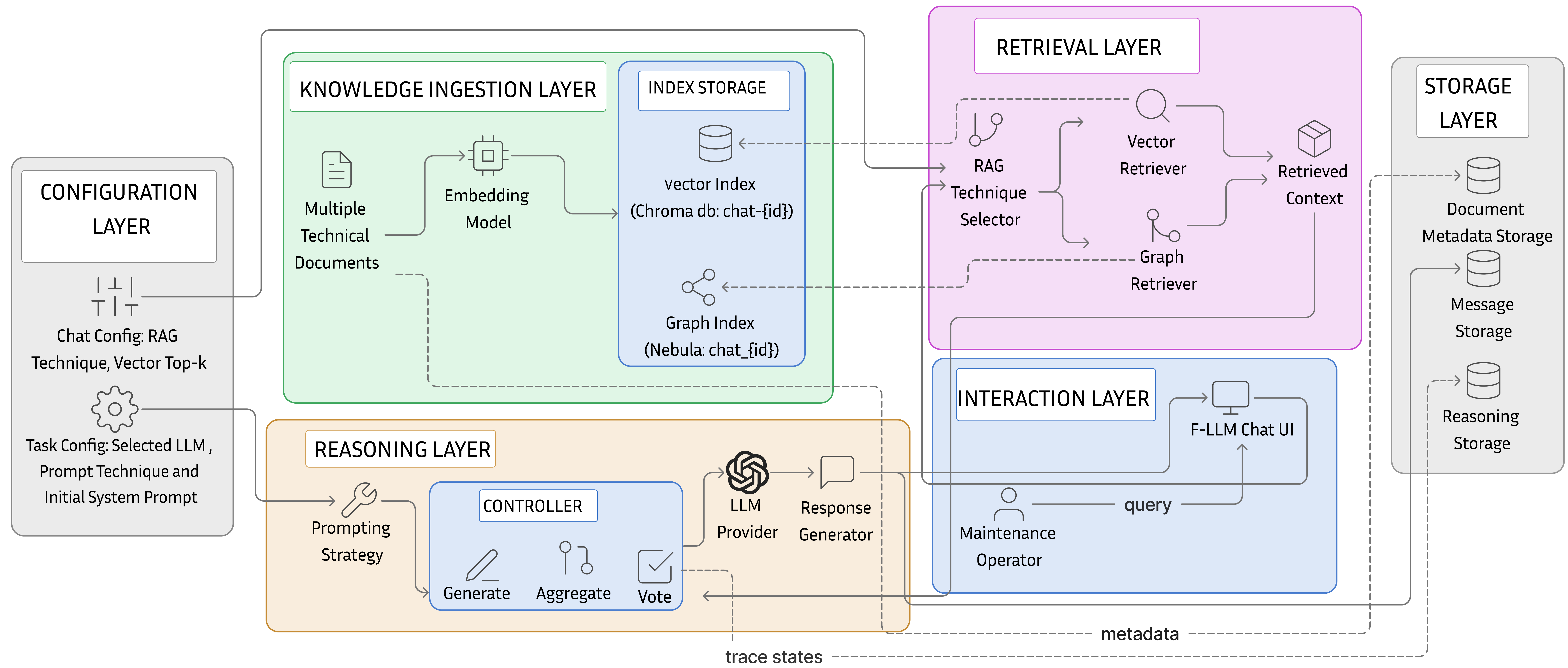}
    \caption{FactoryLLM system architecture. Technical documents from multiple machines are ingested into chat-scoped indices. At query time, the retrieved context set is passed to a configurable reasoning layer that supports IO, CoT, ToT, and GoT prompting strategies before producing a grounded response.}
    \label{fig:architecture}
\end{figure*}

\section{Related Work}
In this section, we explore the current modern efforts on the use of LLMs for improved reasoning and understanding of manufacturing workflows.

\noindent \textbf{\textit{A. LLMs in Manufacturing.}}
The use of Large language models in manufacturing is growing rapidly. Li et al. \cite{li2024large} provide a thorough survey of how LLMs are used in production planning, quality assurance, and maintenance; pointing out that, although LLMs appear to be very good at understanding unstructured data from manufacturing, issues of domain adaptation and dealing with varied data from many different industrial machines still exist. Raza et al. \cite{raza2025industrial} review industrial LLM applications in a number of industries and observe that most implementations are still specific to an industry, only covering one or two domains, and have little ability to generalise across machines. Specifically in maintenance, Getz and Tong \cite{getz2025large} use LLMs to extract information from maintenance logs at a major aerospace manufacturer and discover that LLM performance is greatly reduced by technical jargon, abbreviations and inconsistent formatting, unless the data was carefully prepared. Wang et al. \cite{du2025llm} develop an LLM-based cognitive assistant for aircraft panel drilling, which lowers the cognitive burden on untrained operators by recognising intent and extracting parameters. While these studies show the potential of LLMs for single manufacturing tasks, none of these address the issue of cross-document/domain reasoning across multiple documents from a suite of interconnected machines – the multi-machine scenario that can be evaluated using the proposed FactoryLLM is typical of real-world smart factories.

\noindent \textbf{\textit{B. RAG for Industrial Documentation.}}
Retrieval-Augmented Generation has become the most popular method for grounding LLMs in domain-specific content. Lewis et al. \cite{lewis2020retrieval} introduce the RAG framework, and demonstrate that adding retrieved context to language models greatly improves factual correctness in knowledge-intensive tasks. A number of recent studies have used RAG in manufacturing environments. Kernan Freire et al. \cite{freire2024knowledge} develop a RAG-based knowledge-sharing tool for factory workers at a detergent manufacturer, evaluating seven LLMs and discovering that GPT-4 has the highest factuality while open-source models have comparable performance with lower rates of hallucination. Zhang et al. \cite{zhang2025knowledge} develop a Graph RAG-enhanced maintenance chatbot at Bosch, which processed 49 documents from five production stations, and find that standard vector-based RAG had low precision on maintenance questions, motivating the development of a knowledge-graph-enhanced method. Liu et al. \cite{liu2025knowledge} propose a hybrid RAG technique using BERT-based query classification for industrial question-answering and report significant accuracy gains over basic LLMs. However, each of these systems is designed for a single machine type or a single production setting. None of them do cross-machine retrieval and reasoning – combining responses from the hardware and software documentation of different, interacting devices simultaneously.

\noindent \textbf{\textit{C. Evaluation of RAG Systems.}}
The quality of RAG-generated responses must be assessed, particularly in areas where safety is important, as incorrect advice can have real-world effects. The RAGAS framework \cite{es2024ragas} provides reference-free, LLM-based evaluation using metrics like faithfulness (whether claims are supported by retrieved context), answer relevance (whether the response answers the query), context precision (signal-to-noise ratio of retrieved context), and context recall (whether all needed information was retrieved). ARES \cite{saad2024ares}, uses synthetic data to fine-tune lightweight judges and provides statistical confidence bounds. Roychowdhury et al. \cite{roychowdhury2024evaluation} evaluate RAGAS metrics in the telecom domain and find that automated cosine-similarity-based scoring can be unreliable for specialised technical vocabulary – a finding that is very important to manufacturing, where vocabulary specific to the field is common. Despite the growing number of evaluation tools available, very few studies have used standardised evaluation of LLMs operating in manufacturing maintenance contexts. Most existing research depends on ad hoc question sets and manual inspection, making it difficult to compare findings between studies.

\noindent \textbf{\textit{D. Summary and Research Gaps.}}
The literature shows several recurring gaps. First, current RAG systems for manufacturing are built for single-document or single-machine situations, and disallow cross-machine reasoning across different machines. Second, most implementations are proprietary, limiting reproducibility. Third, no systematic study that tests several LLMs and prompting strategies on manufacturing maintenance tasks using standardised evaluation metrics. FactoryLLM fills all three gaps: it is an open-source, multi-document RAG platform that takes documentation from both an AIV and its Mobile Planner software, allows side-by-side comparison of multiple LLMs and prompting techniques, and uses the RAGAS framework for standardised evaluation – providing a repeatable baseline for cross-machine maintenance reasoning research.

\section{System Overview: FactoryLLM}


This section presents an overview of the FactoryLLM architecture. As illustrated in Figure~\ref{fig:architecture}. FactoryLLM is designed as a six‑layer pipeline that ingests technical documentation generated by multiple machines and produces question‑answering outputs that are well‑substantiated and explicitly traceable to their source documents. Each layer fulfils a distinct functional role, collectively enabling reliable knowledge ingestion, retrieval, reasoning, interaction, and storage. Building on this structure, FactoryLLM adopts a modular, configuration‑driven design paradigm that emphasises reproducibility, transparency, and flexible reasoning over heterogeneous technical sources. Session‑scoped indexing and persistent storage ensure experimental isolation and full traceability, while the clear separation of concerns supports systematic comparison of retrieval and prompting strategies, cross‑machine reasoning, and human‑in‑the‑loop control through a chat‑based interface. The layered architecture is described in detail below.

\textbf{Configuration Layer}: Before processing any user query, this Layer enables operators to define two sets of parameters that govern system behaviour at both the task and interaction levels. At the task level, the operator selects the underlying LLM provider (e.g., OpenAI, OpenRouter, Google Gemini, or locally hosted models), the prompting strategy (including Input–Output, Chain-of-Thought, Tree-of-Thought, or Graph-of-Thought prompting), and an optional initial system message that specifies the role and behavioural constraints of the chat assistant. 

At the interaction (chat) level, the operator configures the RAG strategy as either vector-based or graph-based and specifies the number of top-\(k\) document sections to be retrieved. All configuration settings are persistently stored in the database, thereby ensuring full reproducibility of experiments and enabling consistent evaluation across repeated runs.


\textbf{Knowledge Ingestion Layer}: Following the parameterisation defined in the Configuration Layer, this layer operationalises the document-related settings for a given chat session. Upon session initiation, the operator uploads one or more technical documents in supported formats (PDF, DOCX, and TXT). These documents are converted into  LlamaIndex \texttt{Document} objects and segmented according to the pre-configured \texttt{chunk\_size} and \texttt{chunk\_overlap} settings to enable efficient indexing and retrieval using RAG.

The retrieval strategy selected at the chat level determines the indexing mechanism. For vector-based retrieval, document chunks are embedded and stored in a session-specific ChromaDB collection (\texttt{chat-\{id\}}). For graph-based retrieval, a corresponding session-scoped NebulaGraph index is created. This configuration-driven design ensures reproducible document processing and enables cross-system reasoning by jointly indexing multiple document sources within a single session. For instance, technical documentation from both the AIV and Mobile Planner software can be jointly indexed, allowing the retrieval mechanisms to identify semantic and relational links that span multiple machine subsystems.

\textbf{Interaction Layer} provides the user-facing interface through which queries are submitted and responses are presented. Operators interact with the system via the FactoryLLM chat interface, where a natural‑language query is entered and forwarded to the Retrieval Layer for contextual search. 

Once the query has been processed and a response has been generated by the Reasoning Layer, the Interaction Layer displays the final answer to the user within the same chat interface and therefore serves as the communication bridge between the user and the underlying pipeline, enabling intuitive query submission while transparently presenting grounded, system-generated responses.


\textbf{Retrieval Layer}: When a query is submitted, this Layer selects an appropriate RAG strategy (RAG Technique Selector) based on the configuration defined earlier. In the vector retriever mode, the system performs a similarity-based search over the session-specific document index to identify the top-\(k\) text segments that are most relevant to the query. In the graph retriever mode, the system explores relationships between concepts captured during document ingestion to assemble a relevant set of contextual information.

Regardless of the retrieval strategy used, the output of this layer is a retrieved context, consisting of document sections that are passed to the reasoning layer for answer generation. Because retrieval operates over a unified index built from multiple uploaded documents, a single query can draw supporting information from documentation associated with different machines or software components, enabling cross-system reasoning within the same session.

\textbf{Reasoning Layer} is responsible for generating a grounded response by combining the operator query with the retrieved contextual information. Based on the prompting strategy selected in the Configuration Layer, the system applies an appropriate reasoning approach to guide how the language model produces the final answer. 

Four reasoning strategies are supported:
\begin{itemize}
    \item \textbf{Direct prompting (IO):} the model generates a response in a single step using the provided context.
    \item \textbf{Chain-of-Thought (CoT):} the query is decomposed into a sequence of intermediate reasoning steps, which are processed sequentially before producing the final answer.
    \item \textbf{Tree-of-Thought (ToT):} multiple candidate reasoning paths are explored in parallel, after which the most strongly supported answer is selected.
    \item \textbf{Graph-of-Thought (GoT):} reasoning paths are organised in a graph structure, allowing information from different retrieved passages to be combined before a final decision is made.
\end{itemize}

The strategies that support multi‑path retrieval, the system may generate multiple candidate context sets, aggregate evidence across them, and apply a voting mechanism to select the most consistently supported information for downstream reasoning. The selected reasoning strategy is executed step by step, with the LLM invoked as required at each stage. If no relevant context is available from the Retrieval Layer, the system returns a fallback response rather than generating an unsupported answer. All intermediate reasoning steps are recorded, providing a transparent audit trail that supports traceability and post‑hoc analysis of each generated response.

\textbf{Storage Layer} maintains a complete and structured record of all system interactions and experimental artefacts using a relational database. Each conversation turn, including system prompts, user queries, and generated responses, is stored and linked to its corresponding chat session. Intermediate reasoning states produced during response generation are also recorded, enabling post‑hoc inspection and traceability of how each answer was constructed.

In addition, metadata for all uploaded documents is stored and associated with the relevant session, allowing retrieved content to be accurately linked back to its source files. Collectively, these stored records support reproducibility and systematic evaluation, as ground‑truth contexts, retrieved passages, and generated answers can be reconstructed for any query. This structured storage design enables seamless integration with RAGAS‑based evaluation workflows.

\section{Case Study Setup}\label{sec:eval_methodology}
The evaluation aims to assess how effectively a RAG-augmented LLM can address maintenance queries relating to documentation for two interconnected systems — an Autonomous Intelligent Vehicle (AIV) and its Mobile Planner software. We specifically measure the precision and completeness of retrieved information, and the accuracy of generated responses against the source documentation — characteristics especially important in industrial maintenance, where incorrect advice could cause operational disruptions.

\subsection{Experimental Dataset}\label{sec:dataset}

The evaluation dataset consists of 30 questions (an example presented below) derived from the technical documentation of the AIV and Mobile Planner systems — totalling approximately 600 pages across both manuals. The questions were designed to reflect the types of operational and troubleshooting queries that maintenance operators encounter in the field, covering fault diagnosis, configuration guidance, and procedural instructions. All 30 questions are cross-machine in nature, incorporating concept relationships and requiring information from both the AIV hardware documentation and the Mobile Planner software documentation to be answered fully — no question can be resolved from a single document alone. This design ensures the dataset specifically tests the cross-system reasoning capability that FactoryLLM is built to support. No ground-truth reference answers are provided; instead, evaluation relies on the automated RAGAS and NVIDIA metric frameworks described below, which assess retrieval and generation quality without requiring manually authored reference responses.

\textit{Example Question: I want to add side-mount lasers to my custom payload because the payload is tall. Which physical port on the core do they connect to, and what software parameter handles the data from these specific sensors?}

\subsection{Experimental Configuration}

All experiments are conducted within FactoryLLM with a consistent retrieval setup: a chunk size of 1{,}000 tokens with 200-token overlap, and top-$k = 10$ chunks retrieved per query. This configuration remains fixed across all models to ensure that score variations reflect differences in LLM reasoning capability rather than retrieval settings.

Three LLMs are evaluated under identical conditions:
\begin{itemize}
    \item Qwen3-235B-A22B-Instruct-2507
    \item Llama 4 Maverick
    \item Gemma-3-27B
\end{itemize}

These models were selected to span a range of architectures and parameter scales among openly available LLMs, enabling FactoryLLM to demonstrate its ability to surface performance differences across diverse model families in a cross-machine retrieval setting.

\subsection{Evaluation Metrics}

Response quality is assessed using two complementary frameworks: RAGAS~\cite{es2024ragas} and NVIDIA LLM-as-Judge~\cite{nvidia2024nemotron}. Using dual frameworks addresses a gap in existing work, where most manufacturing RAG evaluations rely on a single metric suite or ad hoc inspection~\cite{roychowdhury2024evaluation}. Together, they cover both retrieval and generation stages at different levels of granularity. All metrics are computed on a $[0, 1]$ scale.

\subsubsection{RAGAS Metrics} RAGAS evaluates retrieval and generation independently through four metrics:
\begin{itemize}
    \item \textbf{Context Precision}: proportion of retrieved chunks relevant to the query (retrieval signal-to-noise).
    \item \textbf{Context Recall}: coverage of the reference answer's required information by the retrieved context.
    \item \textbf{Response Relevancy}: degree to which the generated response directly addresses the query.
    \item \textbf{Faithfulness}: factual consistency of the response with respect to retrieved passages, verifying no unsupported claims are introduced.
\end{itemize}

\subsubsection{NVIDIA LLM-as-Judge Metrics} Two NVIDIA metrics complement RAGAS by providing parallel signals at a coarser granularity:
\begin{itemize}
    \item \textbf{Context Relevance}: pertinence of retrieved context to the query, complementing Context Precision.
    \item \textbf{Response Groundedness}: traceability of response claims to retrieved context, complementing Faithfulness.
\end{itemize}

This pairing enables cross-framework consistency checks — for instance, divergence between RAGAS Faithfulness (strict, claim-level) and NVIDIA Groundedness (holistic) can reveal whether generation issues are localised or systemic.

\subsection{Evaluation Procedure}

For each of the 30 test questions, the query is submitted to FactoryLLM under each of the three LLMs. The platform retrieves the top-10 most similar passages from the combined AIV and Mobile Planner vector index and passes them, along with the query, to the selected LLM. The generated response and retrieved context are then evaluated by both the RAGAS and NVIDIA pipelines to produce per-question scores. Final scores for each LLM are reported as averages across all 30 questions. Detailed results are presented in Section~\ref{sec:exp_results}.

\section{Results and Evaluation}\label{sec:exp_results}

This section presents results from evaluating three LLMs within FactoryLLM across 30 cross-machine questions spanning AIV and Mobile Planner documentation.

\begin{figure*}[t]
\centering
\includegraphics[width=0.7\textwidth]{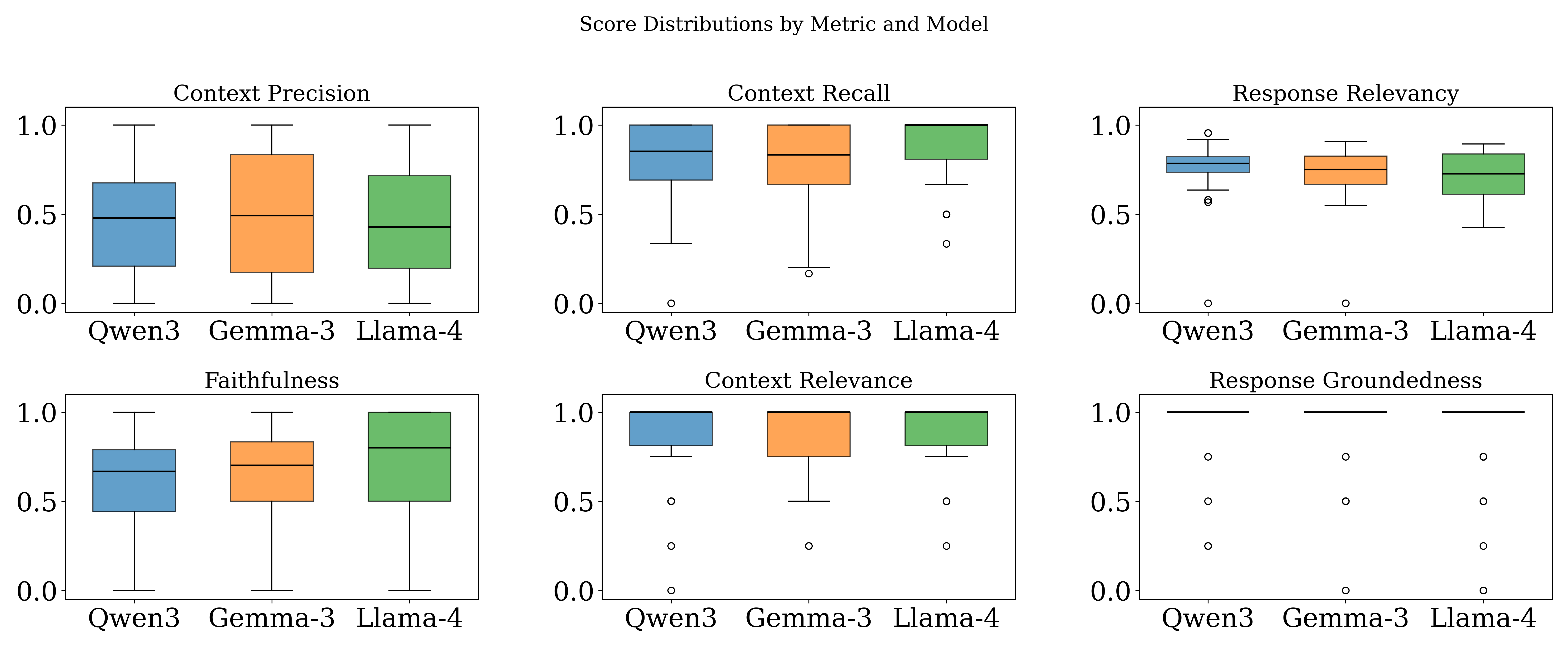}
\caption{Score distributions by metric. Context precision shows the widest spread and lowest median; response groundedness is consistently high.}
\label{fig:boxplots}
\end{figure*}

(Fig.~\ref{fig:boxplots}) summarises mean scores across all metrics. All models achieve overall averages between 0.73 and 0.76, confirming that RAG-based cross-machine reasoning is feasible within FactoryLLM. As shown in Fig.~\ref{fig:boxplots}, retrieval-side metrics (context precision: 0.46--0.51) are consistently weaker than generation-side metrics (groundedness: 0.88--0.95), revealing that the primary bottleneck lies in retrieval rather than generation — a diagnosis enabled by FactoryLLM's separation of evaluation stages~\cite{zhang2025knowledge, freire2024knowledge}.

Context precision is the weakest metric (0.46--0.51): roughly half of retrieved passages are not directly useful, as cross-machine queries pull topically adjacent but irrelevant chunks from both document sets. Six questions exhibit mean precision below 0.20. Context recall is substantially higher (0.76--0.89), meaning relevant information is present but diluted by noise (Fig.~\ref{fig:boxplots}). This precision--recall gap points to re-ranking and query decomposition as clear optimisation targets. To illustrate cross-machine reasoning, we use the example query from Section~\ref{sec:dataset}, which asks how to add side-mount lasers to a custom payload. Running Qwen3-235B over the AIV user manual and the Mobile Planner manual, FactoryLLM draws the RS-232 Aux Sensors connector (DB9 female port) from the AIV hardware documentation and the Safety Commissioning procedure (Main Menu \textgreater{} Robot \textgreater{} Safety Commissioning) from the Mobile Planner manual. Neither manual answers the query alone. By combining the hardware connection details with the software commissioning steps, FactoryLLM produces a single grounded response that spans both sources. NVIDIA context relevance (0.86--0.90) confirms content is broadly on-topic but lacks the specificity that RAGAS precision measures — illustrating the value of dual-framework evaluation.

\begin{figure}[!h]
\centering
\includegraphics[width=\columnwidth]{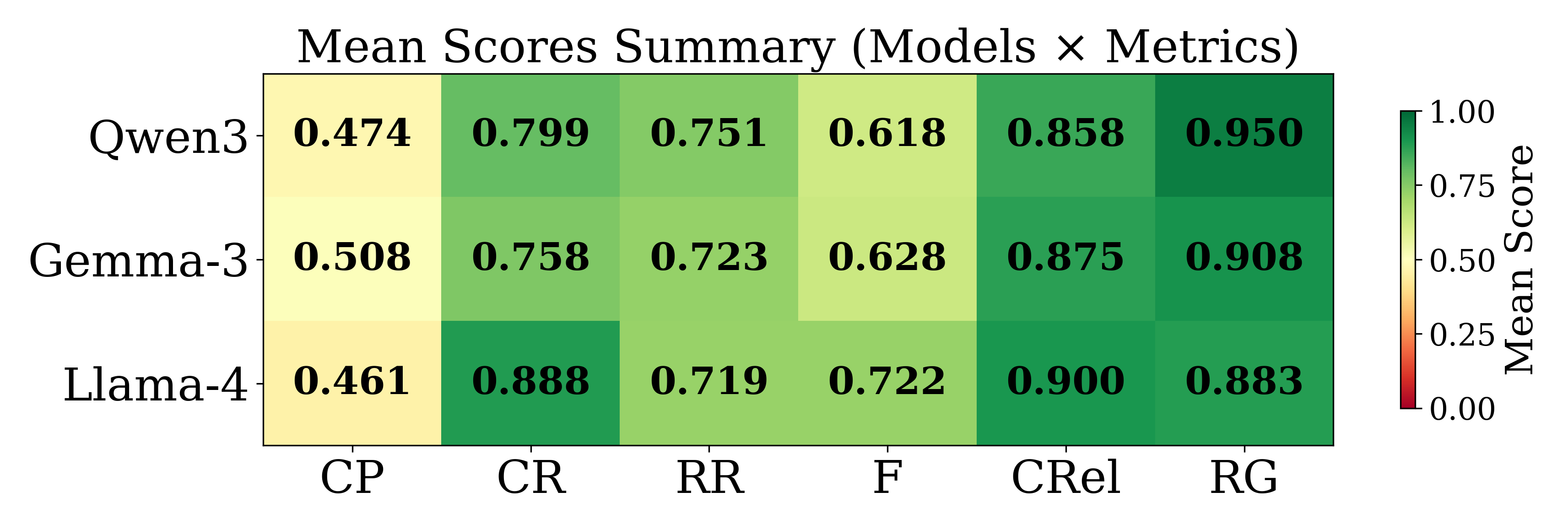}
\caption{Mean scores across models and metrics. Retrieval metrics (left) are lower and more variable than generation metrics (right).}
\label{fig:heatmap}
\end{figure}

Response groundedness (0.88--0.95) confirms FactoryLLM's RAG pipeline effectively prevents hallucination, contrasting with concerns in prior work~\cite{freire2024knowledge}. Faithfulness varies more (0.62--0.72), with seven questions below 0.50, indicating certain cross-machine queries produce partially unsupported claims. FactoryLLM surfaces these at per-question granularity (Fig.~\ref{fig:heatmap}), enabling targeted diagnosis. Response relevancy is moderate (0.72--0.75).

Two findings emerge: (1)~cross-machine RAG is feasible but retrieval-limited — groundedness exceeds 0.88 while precision averages 0.48; (2)~RAGAS and NVIDIA metrics provide complementary signals at different strictness levels ($r=0.61$ between recall and faithfulness; groundedness averages 0.91 vs.\ faithfulness 0.66), enabling robust multi-faceted assessment.

\section{Conclusion}
This paper introduces FactoryLLM, a safe and open LLM playground for evaluating the reasoning abilities of RAG-based LLMs when dealing with documentation from multiple interconnected industrial machines. FactoryLLM ensures safety by supporting local and open-source LLMs, allowing experimentation without exposing sensitive industrial data, and provides a controlled environment for systematic evaluation. The Playground is open and publicly available, enabling researchers and practitioners to plug in different LLMs and replicate or extend experiments. Through a case study involving an AIV and its Mobile Planner software, we evaluated three LLMs across 30 cross-machine queries. The results confirm that cross-machine RAG is feasible, achieving high groundedness despite limited retrieval precision, while the dual RAGAS and NVIDIA frameworks support robust assessment of both retrieval and generation. Future work will incorporate human expert evaluation and scale FactoryLLM to additional document types.

\section*{Acknowledgment}
Parts of this research was supported by the Australian Research Council (ARC) Industrial Transformation Research Hub for Future Digital Manufacturing (DMH), IH230100013

\bibliographystyle{IEEEtran}
\bibliography{references}

@article{lu2017industry,
  title={Industry 4.0: A survey on technologies, applications and open research issues},
  author={Lu, Yang},
  journal={Journal of industrial information integration},
  volume={6},
  pages={1--10},
  year={2017},
  publisher={Elsevier}
}

@article{siemens2024true,
  title={The true cost of downtime 2024.},
  author={Siemens, AG},
  journal={Siemens AG},
  year={2024}
}

@inproceedings{wan2014knowledge,
  title={Knowledge management for maintenance, repair and service of manufacturing system},
  author={Wan, Shan and Gao, JX and Li, Dongbo and Evans, RD},
  booktitle={International Conference on Manufacturing Research},
  year={2014},
  organization={Southampton Solent University}
}

@book{osti_5675197,
  author       = {Jackson, P},
  title        = {Introduction to expert systems},
  url          = {https://www.osti.gov/biblio/5675197},
  place        = {United States},
  publisher    = {Addison-Wesley Pub. Co.,Reading, MA},
  year         = {1986},
  month        = {01}}

@inproceedings{zhang2025knowledge,
  title={Knowledge Graph Construction towards a Graph RAG-Enhanced Intelligent Maintenance Chatbot},
  author={Zhang, Hansi and Schmidt, Wilma Johanna and Shen, Xiaozhi and Cao, Qiushi and Monka, Sebastian and Paschke, Adrian},
  booktitle={International Workshop on Scaling Knowledge Graphs for Industry 2025},
  year={2025}
}

@article{raza2025industrial,
  title={Industrial applications of large language models},
  author={Raza, Mubashar and Jahangir, Zarmina and Riaz, Muhammad Bilal and Saeed, Muhammad Jasim and Sattar, Muhammad Awais},
  journal={Scientific Reports},
  volume={15},
  number={1},
  pages={13755},
  year={2025},
  publisher={Nature Publishing Group UK London}
}

@article{li2024large,
  title={Large language models for manufacturing},
  author={Li, Yiwei and Zhao, Huaqin and Jiang, Hanqi and Pan, Yi and Liu, Zhengliang and Wu, Zihao and Shu, Peng and Tian, Jie and Yang, Tianze and Xu, Shaochen and others},
  journal={arXiv preprint arXiv:2410.21418},
  year={2024}
}

@article{freire2024knowledge,
  title={Knowledge sharing in manufacturing using large language models: User evaluation and model benchmarking},
  author={Freire, Samuel Kernan and Wang, Chaofan and Foosherian, Mina and Wellsandt, Stefan and Ruiz-Arenas, Santiago and Niforatos, Evangelos},
  journal={arXiv preprint arXiv:2401.05200},
  year={2024}
}

@article{lewis2020retrieval,
  title={Retrieval-augmented generation for knowledge-intensive nlp tasks},
  author={Lewis, Patrick and Perez, Ethan and Piktus, Aleksandra and Petroni, Fabio and Karpukhin, Vladimir and Goyal, Naman and K{\"u}ttler, Heinrich and Lewis, Mike and Yih, Wen-tau and Rockt{\"a}schel, Tim and others},
  journal={Advances in neural information processing systems},
  volume={33},
  pages={9459--9474},
  year={2020}
}

@article{liu2025knowledge,
  title={Knowledge Enhanced Industrial Question-Answering Using Large Language Models},
  author={Liu, Ronghui and Ren, Hao and Ren, Haojie and Rui, Wu and Cui, Wei and Liang, Xiaojun and Yang, Chunhua and Gui, Weihua},
  journal={Engineering},
  year={2025},
  publisher={Elsevier}
}

@inproceedings{getz2025large,
  title={Large Language Model Accelerated Maintenance Insights},
  author={Getz, Noah and Tong, Xiaorui},
  booktitle={Annual Conference of the PHM Society},
  volume={17},
  number={1},
  year={2025}
}

@article{du2025llm,
  title={LLM-MANUF: An integrated framework of Fine-Tuning large language models for intelligent Decision-Making in manufacturing},
  author={Du, Kaze and Yang, Bo and Xie, Keqiang and Dong, Nan and Zhang, Zhengping and Wang, Shilong and Mo, Fan},
  journal={Advanced Engineering Informatics},
  volume={65},
  pages={103263},
  year={2025},
  publisher={Elsevier}
}

@inproceedings{es2024ragas,
  title={Ragas: Automated evaluation of retrieval augmented generation},
  author={Es, Shahul and James, Jithin and Anke, Luis Espinosa and Schockaert, Steven},
  booktitle={Proceedings of the 18th Conference of the European Chapter of the Association for Computational Linguistics: System Demonstrations},
  pages={150--158},
  year={2024}
}

@inproceedings{saad2024ares,
  title={Ares: An automated evaluation framework for retrieval-augmented generation systems},
  author={Saad-Falcon, Jon and Khattab, Omar and Potts, Christopher and Zaharia, Matei},
  booktitle={Proceedings of the 2024 Conference of the North American Chapter of the Association for Computational Linguistics: Human Language Technologies (Volume 1: Long Papers)},
  pages={338--354},
  year={2024}
}

@article{roychowdhury2024evaluation,
  title={Evaluation of rag metrics for question answering in the telecom domain},
  author={Roychowdhury, Sujoy and Soman, Sumit and Ranjani, HG and Gunda, Neeraj and Chhabra, Vansh and Bala, Sai Krishna},
  journal={arXiv preprint arXiv:2407.12873},
  year={2024}
}

@misc{nvidia2024nemotron,
      title={Nemotron-4 340B Technical Report}, 
      author={Nvidia and Bo Adler and Niket Agarwal et al.},
      year={2024},
      eprint={2406.11704},
      archivePrefix={arXiv},
      primaryClass={cs.CL},
      url={https://arxiv.org/abs/2406.11704}, 
}

\end{document}